# Applying the Wizard-of-Oz Technique to Multimodal Human-Robot Dialogue

Matthew Marge, Claire Bonial, Brendan Byrne, Taylor Cassidy,
A. William Evans, Susan G. Hill, and Clare Voss

*Abstract*— Our overall program objective is to provide more natural ways for soldiers to interact and communicate with robots, much like how soldiers communicate with other soldiers today. We describe how the Wizard-of-Oz (WOz) method can be applied to multimodal human-robot dialogue in a collaborative exploration task. While the WOz method can help design robot behaviors, traditional approaches place the burden of decisions on a single wizard. In this work, we consider two wizards to stand in for robot navigation and dialogue management software components. The scenario used to elicit data is one in which a human-robot team is tasked with exploring an unknown environment: a human gives verbal instructions from a remote location and the robot follows them, clarifying possible misunderstandings as needed via dialogue. We found the division of labor between wizards to be workable, which holds promise for future software development.

## I. INTRODUCTION

To enable effective teaming with humans, there is a need for improved human-robot communication, particularly two-way spoken dialogue. Natural language dialogue has the benefit of familiarity and flexibility for humans, allowing information such as tasking, situational awareness, and goals to be exchanged. The long-term objective of this work is to have robots operate as intelligent agents that are teammates with humans. To this end, we are aiming for as natural communication as possible between robots and humans. In this work, *natural communication* must meet several requirements. The communication must be *situated* so that it refers to the environment, build upon references to physical common ground, and must be initiated at appropriate times.

In the near-term, our objective is to investigate how humans think natural communication with a robot should work. We are conducting studies with human participants that elicit natural language communication with a robot. We employ the Wizard-of-Oz method [1], in which humans simulate the robot's actions without the participant's awareness. We hypothesize that an effective separation point for robot intelligence is between its multimodal dialogue management (text, speech, LIDAR mapping, images, and video) and autonomous navigation capabilities. We opted to use two wizards in our setup, so only the wizard managing the robot's communications intelligence has bidirectional communications with the participant.

M. Marge, C. Bonial, B. Byrne, T. Cassidy, A. W. Evans, S. G. Hill, and C. Voss are with the U. S. Army Research Laboratory, Adelphi, MD 20783, USA. Corresponding author email: matthew.r.marge.civ@mail.mil

## II. BACKGROUND

Wizard-of-Oz (WOz) design is a useful tool for HRI researchers and has been widely used within the HRI research literature (e.g., [2][3]). The WOz design is appealing to HRI researchers because it allows for low development cost and extremely malleable system functionality [4]. Researchers using WOz techniques are able to simulate capabilities for autonomous systems that are not fully developed or do not yet exist, through the use of confederates substituting for autonomous systems intelligence. This allows HRI researchers to investigate human-system integration issues early before investing large amounts of time and development into system capabilities.

We used the two-wizard method to separate out dialogue behaviors from robot navigation, extending our previous pilot trials that used a single wizard [5]. With this setup, we hope to gain a better, more systematic understanding of effective dialogue and turn-taking. Others (e.g., [6]) have considered the multi-wizard setup for multimodal interfaces. Our work seeks to expand on this method by addressing multimodal communication when the robot and human are not co-present – where information like robot position, visual media, and dialogue would need to be exchanged.

## III. APPROACH

The goal of these experiments is to collect natural language communications between robots and humans. Below we describe the task and setup.

*A. Collaborative Exploration Task*

The participant's task is to instruct a robot in a remote location to navigate through an indoor environment. The participant's understanding of the environment is limited to an interface (see Figure 1) and dialogue interactions with the robot. The participant does not have access to live video from the robot, and must make verbal requests for images instead. Participants provide unconstrained instructions (e.g., "turn left here", "move forward five feet") as well as respond to the robot's clarification questions.

Participants interact with a customized Clearpath Robotics Jackal robot that consists of a PrimeSense RGB camera, an inertial measurement unit, and a laser scanner.

*B. Two Wizard Setup*

We developed a two-wizard setup that aims to divide multimodal processing labor between the *robot navigator* (RN) and the *dialogue manager* (DM) wizards. The RN acts as a surrogate for autonomous navigation algorithms, using a rich collection of raw and partially processed sensor data. The DM acts as the natural language communications intelligence of the robot, translating verbal commands given by the participant into movement orders in constrained natural language that are issued to the RN. To illustrate the general experimental framework, Figure 1 shows the interactions between the participant and wizards. The participant communicates exclusively with the DM, who in turn communicates with both the participant and RN.

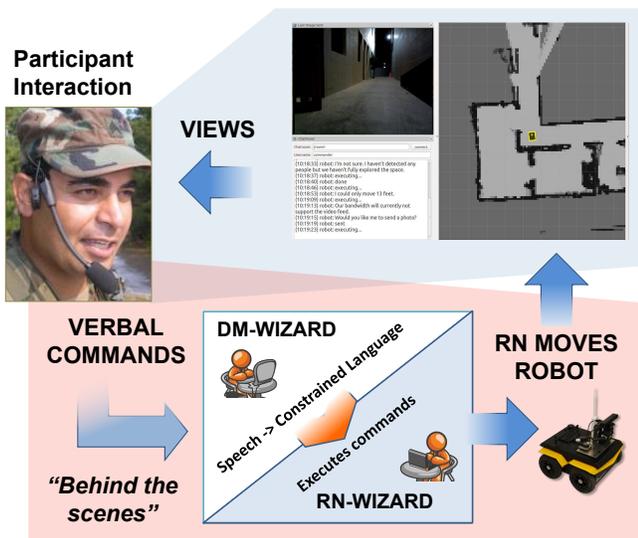

Fig. 1. The participant provides instructions, unaware that the robot is composed of two wizards. The DM wizard translates spoken language commands into a constrained action list for the RN wizard to execute. Map updates, image data, and the robot's dialogue responses are sent to the participant through a multimodal interface (*upper right*).

The participant sits at a workstation that displays information from the robot. An interface shows a 2-dimensional representation of the robot's environment, the robot's current location and orientation, the chat window with the robot, and the last image that the robot sent to the participant. The wizards use a second workstation area, where the DM and RN sit next to each other. The DM relays communications between the participant and RN using two separate chat windows – one for communications from the DM to the participant, and one from the DM to the RN. Note that these are unidirectional chat windows, the DM receives input from the participant and RN as speech.

As part of the experiment development process, we designed *guidelines* governing the DM's use of clarification questions with the participant. They consist of dialogue strategies for common situations, as well as decisions on what the robot should consider problematic or executable. The DM guidelines were iteratively refined during a piloting phase to support conversation in a broad set of contexts.

### C. Instrumentation

The participant and RN use microphones to interact with the DM. The DM listens with headphones via a VoIP server.

## IV. DISCUSSION

Our methodology adopted a division of labor in which the DM was capable of grounding and interpreting language in the context of the robot's physical surroundings (i.e., *situated reasoning*). Accordingly, the DM wizard interface was augmented with the same LIDAR map viewed by the participant, any images requested and sent to the participant, as well as the video feed seen by the RN. With the addition of these sources of information, the DM was equipped to manage a variety of communication issues. In general, the DM's responsibilities were expanded from addressing only linguistically problematic instructions to addressing any problematic mismatches between the instructions and the environment that could be recognized given the DM's sources of information. As a result, most problematic instructions are filtered out by the DM, and only linguistically well-formed, physically appropriate instructions are passed to the RN. Thus, the RN's communication responsibilities are narrowed to addressing cases where the DM failed to see the problematic nature of an instruction given the physical environment, or cases where the RN's expertise in the precise movements and capabilities of the robot are necessary for understanding if an instruction can be carried out or not.

## V. SUMMARY

We provide a methodology for studying multimodal human-robot dialogue, along with principles for the design of Wizard-of-Oz (WOz) experimentation. Traditional WOz methods can burden a single wizard with multiple, compounded layers of intelligence. A two-wizard setup can isolate natural language dialogue behaviors from robot navigation. In future work, we will constrain some of the wizard behaviors through a specialized interface, with the goal of collecting training data for autonomous dialogue processing.

## ACKNOWLEDGMENTS

The authors thank David Traum, Ron Artstein, David Baran, Reginald Hobbs, and Douglas Summers-Stay for their contributions to this project.